\documentclass[11pt]{article}

\usepackage[preprint]{acl}

\usepackage{times}
\usepackage{latexsym}

\usepackage[T1]{fontenc}
\usepackage{textcomp}
\usepackage[utf8]{inputenc}

\usepackage{microtype}

\usepackage{inconsolata}

\usepackage{graphicx}

\usepackage{subcaption}
\usepackage{booktabs}
\usepackage{multirow}

\usepackage{amsmath}
\usepackage{amssymb}
\usepackage{mathtools}
\usepackage{amsthm}

\usepackage[capitalize,noabbrev]{cleveref}
\usepackage{algorithm}
\usepackage{xcolor}
\usepackage{listings}
\usepackage{xurl}
\usepackage{changepage}
\usepackage{inconsolata}

\definecolor{commentcolor}{RGB}{0, 128, 128} % Teal 
\definecolor{keywordcolor}{RGB}{0, 0, 255} % Blue

\definecolor{codeblue}{RGB}{0, 0, 255}
\definecolor{codegray}{RGB}{160, 160, 160}
\definecolor{codepurple}{RGB}{158, 0, 255}
\definecolor{backcolour}{RGB}{255, 255, 255}

\lstdefinestyle{mystyle}{
    backgroundcolor=\color{backcolour},
    commentstyle=\color{codegray},
    keywordstyle=\color{codeblue},
    numberstyle=\tiny\color{codegray},
    stringstyle=\color{codepurple},
    basicstyle=\ttfamily\footnotesize,
    breakatwhitespace=false,
    breaklines=true,
    captionpos=b,
    keepspaces=true,
    numbers=none,
    showspaces=false,
    showstringspaces=false,
    showtabs=false,
    tabsize=2,
    frame=none,
    lineskip=-1pt,
    otherkeywords={self},
}

\usepackage[textsize=tiny]{todonotes}
\usepackage[dvipsnames]{xcolor}

\title{DECO: Sparse Mixture-of-Experts\\with Dense-Comparable Performance on End-Side Devices}

\author{
 \textbf{Chenyang Song},
 \textbf{Weilin Zhao},
 \textbf{Xu Han},
 \textbf{Chaojun Xiao},
 \textbf{Yingfa Chen},
 \textbf{Zhiyuan Liu}
\\
\\
 Dept. of Comp. Sci. \& Tech., Institute for AI, Tsinghua University, Beijing, China
\\
 {\tt scy22@mails.tsinghua.edu.cn, \{han-xu,liuzy\}@tsinghua.edu.cn}
}

\begin{document}
\maketitle
\begin{abstract}
While Mixture-of-Experts (MoE) scales model capacity without proportionally increasing computation, its massive total parameter footprint creates significant storage and memory-access bottlenecks, which hinder efficient end-side deployment that simultaneously requires high performance, low computational cost, and small storage overhead.
To achieve these properties, we present \textbf{DECO}, a sparse MoE architecture designed to match the performance of dense Transformers under identical total parameter budgets and training tokens.
DECO utilizes the differentiable and flexible ReLU-based routing enhanced by learnable expert-wise scaling, which adaptively balances the contributions of routed and shared experts.
Furthermore, we introduce NormSiLU, an activation function that normalizes inputs prior to SiLU operators, producing a more stable trend of routed-expert activation ratio and a higher intrinsic sparsity level.
We also identify an empirical advantage in using non-gated MLP experts with ReLU-based routing, indicating the possibility of MoE architecture simplification.
Experiments demonstrate that DECO, activating only 20\% of routed experts, matches dense performance and outperforms established MoE baselines.
Our specialized acceleration kernel delivers a 2.93$\times$ speedup on Jetson AGX Orin compared with dense inference. Code and checkpoints are available at \url{https://github.com/thunlp/DECO}.
\end{abstract}

\section{Introduction}

The scale of large language models (LLMs) has grown rapidly to achieve consistent performance gains across diverse tasks. The rising training and deployment costs for massive LLMs have made mixture-of-experts (MoE) an increasingly prominent model architecture. The key property of MoE is the sparse activation, namely, activating a small subset of expert modules from a large pool of parameters. Therefore, MoE retains high capacity and strong performance while substantially reducing computation costs.

As a research hotspot, MoE has been extensively studied, from architecture design~\cite{liu2024deepseekv3,cai2025survey} to scaling laws and compute-optimal settings~\cite{krajewski2024scaling,tian2025towards}. Prior work primarily pursues two objectives: \textbf{high performance} and \textbf{low computation cost}. However, when it comes to end-side deployment, a third non-negligible objective emerges: \textbf{small storage overhead}. Concretely, MoE with a huge number of total parameters demands substantial storage space. More critically, large MoE models may incur high memory-access costs when transferring experts between GPU high-bandwidth memory and shared memory~\cite{li2025can}, or when moving offloaded parameters from disk/flash storage to GPU/NPU memory of end-side devices. Such latency can erode the efficiency gains afforded by sparse computation.

\begin{figure}[t]
    \centering
    \includegraphics[width=0.95\linewidth]{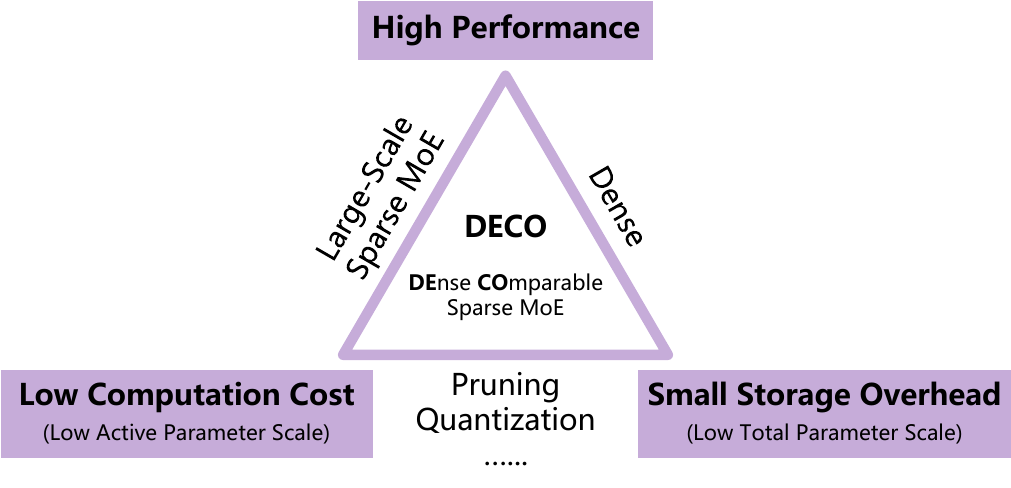}
    \caption{The ``ideal triangle'' of end-side MoE. Beyond the high performance and reduced computational cost of sparse MoE, the model should maintain a minimal storage footprint, achieving high performance within dense-comparable total parameter budgets.}
    \label{fig:ideal-triangle}
    \vspace{-1.5em}
\end{figure}

\begin{figure*}[t]
    \centering
    \includegraphics[width=0.95\linewidth]{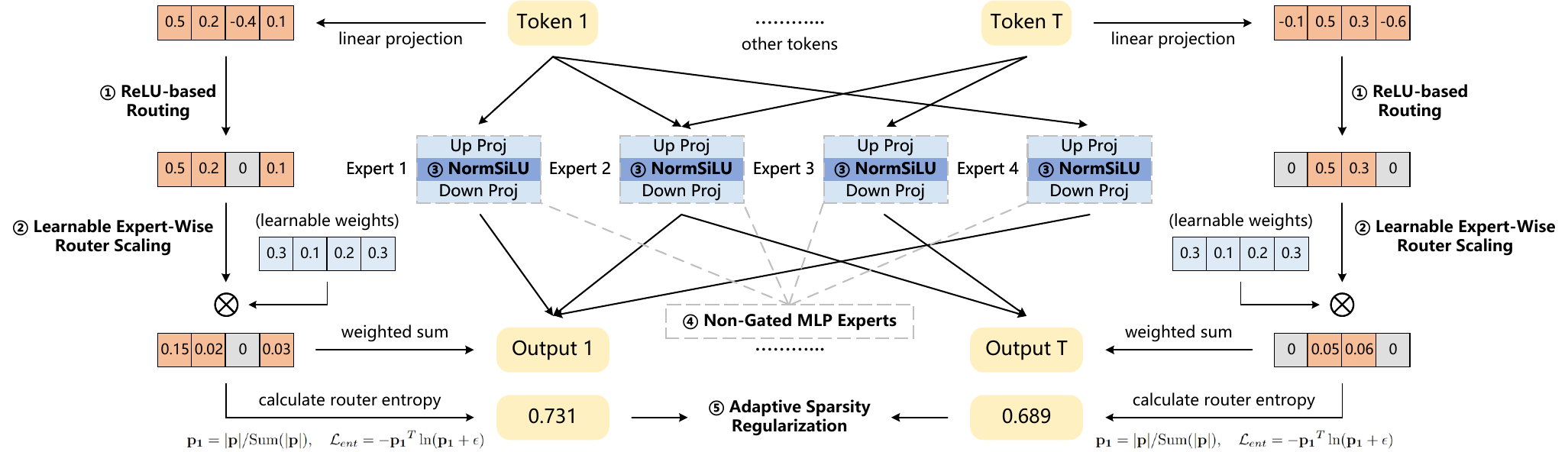}
    \caption{The overall architecture of DECO. For router design, we adopt ReLU-based routing enhanced by learnable expert-wise router scaling. For expert design, we propose NormSiLU as a better routed-expert activation function and employ non-gated MLP experts. For precise sparsity control, we employ adaptive sparsity regularization.}
    \label{fig:deco-main}
    \vspace{-1.3em}
\end{figure*}

Therefore, as shown in Figure~\ref{fig:ideal-triangle}, an ideal MoE model for end-side deployment should satisfy the above three objectives. To pursue this ``ideal triangle'', we pose the question:

\begin{adjustwidth}{1em}{1em}
\textit{Can a sparse MoE model achieve performance comparable to a dense model, given the same total parameter budget and the same number of training tokens?}
\end{adjustwidth}

A closely related study by~\citet{li2025can} identifies the optimal settings of DeepSeek-V3-style MoE architectures that enable them to surpass dense models under matched total parameters and computation budget. However, due to the low per-token computation of sparse MoE, in that work, MoE settings are trained on substantially more tokens under the same computation budget. We adopt a stricter setting that requires exactly the same number of training tokens.

To achieve this goal, we propose \textbf{DECO} (Figure~\ref{fig:deco-main}), a sparse MoE architecture that achieves \textbf{DE}nse-\textbf{CO}mparable performance through a fundamental revision of MoE design.

For router design, conventional MoE models generally adopt TopK routing, which is non-differentiable and enforces a uniform activation ratio across all tokens. To overcome this issue, we adopt \textbf{ReLU-based routing}, a differentiable paradigm that enables flexible token-dependent activation ratios. Moreover, to mitigate output scale imbalances between shared and routed experts, while simultaneously accounting for potential expert heterogeneity, we introduce \textbf{learnable expert-wise router scaling}. This mechanism involves learnable scaling factors to calibrate the contribution of individual routed experts.

The expert design is similarly optimized for stability and efficiency. Empirical analysis reveals that coupling ReLU-based routing with vanilla SiLU-activated experts results in two critical problems: a surging routed-expert activation ratio (Figure~\ref{fig:normsilu-act}) and vanishing SiLU output magnitudes (Figure~\ref{fig:normsilu-magnitude}). To resolve these issues, we propose \textbf{NormSiLU}, which applies dual-stage normalization prior to the SiLU operator. NormSiLU stabilizes activation trends and reduces the activation ratio, alleviating the need for aggressive sparsity regularization. It also produces more stable and significant SiLU output magnitudes, promoting better utilization of expert parameters.
Beyond the activation function, we employ \textbf{non-gated MLP experts} rather than standard gated variants, as they exhibit superior empirical compatibility with ReLU-based routing. Finally, to precisely control the activation ratio, we design an \textbf{adaptive sparsity regularization} that auto-scales the regularization strength.

As shown in Section~\ref{sec:experiments}, at the activation ratio of 20\%, DECO demonstrates performance comparable to dense models when matched for total parameters and training tokens. DECO also surpasses established MoE baselines of the same scale and activation ratio.

% As shown in Section~\ref{sec:experiments}, DECO demonstrates performance comparable to dense models when matched for total parameters and training tokens. DECO also surpasses established MoE baselines of the same scale and activation ratio. Naturally, DECO's dense-comparability is conditional, since the performance of MoE is affected by many factors, including the activation ratio, expert granularity, and shared expert size. In Section~\ref{sec:key-factor}, we analyze the effect of these factors.

Finally, we implement a tailored acceleration kernel for DECO to test its practical inference acceleration value. Based on CUTLASS~\cite{Thakkar_CUTLASS_2023}, the kernel leverages tensor cores to improve computational throughput and reduces memory-access overhead by exploiting the sparse activation. Overall, the kernel achieves a speedup of 2.93$\times$ compared with vanilla dense inference on the real edge device of Jetson AGX Orin.

\section{Related Works}

To achieve high performance while curbing computational growth, MoE has recently risen as the mainstream architecture. An MoE typically comprises three components: a router, a set of experts, and an auxiliary training objective.

\textbf{Router design.} The router computes weights assigned to each expert and selects which experts to activate. It generally consists of a linear projection, an activation function, and post-processing of router scores.

The activation function controls the expert selection pattern. Many MoE designs use TopK, which forces each token to activate a fixed number of experts~\cite{jiang2024mixtral,dai2024deepseekmoe}. However, TopK is criticized for its inflexibility (an input-invariant number of active experts) and non-differentiability. TopP~\cite{huang2024harder} selects experts by a threshold $p$, activating experts until their cumulative router score reaches at least $p$, thereby permitting token-dependent activation ratios. MoE++~\cite{jin2024moe++} retains TopK but introduces zero-computation experts, which indirectly allows variable computation cost. To improve differentiability, ReMoE~\cite{wang2024remoe} and BlockFFN~\cite{song2025blockffn} adopt ReLU for expert selection. Since ReLU naturally produces considerable zero outputs while remaining differentiable, it enables smoothly learnable activation ratios and delivers performance advantages.

Post-processing of router scores primarily normalizes expert weights to preserve a consistent output scale. Most designs use Softmax as the score normalizer. DeepSeek-V3~\cite{liu2024deepseekv3} instead applies element-wise Sigmoid followed by unit-sum normalization, which helps mitigate extremely skewed score distributions. Notably, DeepSeek-V3 also introduces a scalar scaling factor applied to router scores, which helps balance contributions between shared and routed experts.

In this work, DECO adopts ReLU-based routing, and replaces the fixed scalar scaling factor with learnable expert-wise router scaling factors, providing flexibility and accommodating potential heterogeneity in expert output scales.

\textbf{Expert design.} In most mainstream MoE models, each expert is a standard gated MLP with SiLU activation (SwiGLU)~\cite{shazeer2020glu}. DeepSeekMoE~\cite{dai2024deepseekmoe} shows the benefits of introducing fine-grained experts and shared experts but retains the SwiGLU backbone.

DECO refines expert design by introducing NormSiLU as the expert activation, which resolves the issues of surging routed-expert activation ratio and vanishing SiLU output magnitudes. Moreover, we find that with ReLU-based routing, non-gated MLP experts empirically bring a more stable trend of activation ratio than the gated variant.

\textbf{Auxiliary training objective.} Aside from the language modeling loss, MoE models generally introduce auxiliary training objectives. The most common one is for load balancing, typically implemented via the auxiliary loss proposed by~\citet{fedus2022switch}. To alleviate auxiliary-loss interference with language modeling, DeepSeek-V3 adopts a loss-free load-balancing policy without a differentiable objective~\cite{wang2024auxiliary}. MoE models with variable activation ratios often incorporate a sparsification objective. For example, ReMoE applies adaptive L1-norm regularization, and BlockFFN employs chunk-wise sparsification~\cite{wang2024remoe,song2025blockffn}.

Inspired by ReMoE, DECO uses an adaptive sparsity regularization, whose coefficient auto-scales to precisely control the sparsity level. We also replace the L1-norm with router entropy to improve numerical stability.

\section{Methodology of DECO} \label{sec:methodology}

We propose DECO, a sparse MoE architecture with performance comparable to dense variants, given the same total number of parameters and training tokens. We split our design into three components: the router (Section~\ref{sec:router-design}), experts (Section~\ref{sec:expert-design}), and adaptive sparsity regularization (Section~\ref{sec:adaptive-reg}).

\subsection{Router Design} \label{sec:router-design}

\textbf{ReLU-based routing.} Distinct from conventional TopK routing, DECO incorporates ReLU in the router to determine expert activation. As demonstrated in prior studies~\cite{yao2025densemixer,song2025blockffn}, ReLU is fully differentiable, inherently induces sparsity, and supports token-dependent activation ratios. These attributes render ReLU a robust and flexible routing function.

\textbf{Learnable expert-wise router scaling.} To balance the output scales of routed and shared experts, DECO applies a scaling operator to the routing scores before they are multiplied by the expert outputs. We extend the fixed scalar scaling factor of DeepSeek-V3~\cite{liu2024deepseekv3} to a learnable vectorized one. This modification accommodates the potential heterogeneity across routed experts by assigning them distinct, learnable coefficients.

Formally, given the hidden dimension $d_h$, the expert number $N_e$, and the input hidden state $\mathbf{x}\in\mathbb{R}^{d_h}$, the router score $\mathbf{p}$ can be computed as follows:
\begin{equation}
\label{eq:router-design}
    \mathbf{p}=\boldsymbol{\alpha}\odot\mathrm{ReLU}(\mathbf{W}_{router}^T\mathbf{x}),
\end{equation}
where $\mathbf{W}_{router}\in\mathbb{R}^{d_h\times N_e}$ and $\boldsymbol{\alpha}\in\mathbb{R}^{N_e}$ are learnable weights, and $\odot$ is element-wise multiplication.

\begin{algorithm*}[ht]
\caption{Pseudocode of NormSiLU.}
\label{alg:norm_silu}
% \lstset{
%     language=Python,
%     basicstyle=\ttfamily\footnotesize,
%     lineskip=-1pt,
%     keywordstyle=\color{keywordcolor}\bfseries,
%     commentstyle=\color{commentcolor},
%     stringstyle=\color{red},
%     showstringspaces=false,
%     breaklines=true,
%     escapeinside={(*}{*)}
% }
\begin{lstlisting}[language=Python]
# Nt: num_tokens; De: dim_expert; Dh: dim_hidden
# Ne: num_experts; Na: num_active_experts
rms_norm = RMSNorm(dim=dim_expert)  # learnable RMSNorm layer, weight shape is [De]

def NormSiLU(input_hidden, up_proj_weight, intermediate_state):
    # input_hidden: [Nt, Dh], the input hidden state of MoE
    # up_proj_weight: [Ne, De, Dh], the up-projection weights of experts
    # intermediate_state: [Nt, Na, De], the expert intermediate state produced by a sparse linear operation between "input_hidden" and "up_proj_weight"

    # inter-expert mean normalization: conducted at the "num_experts" dimension
    up_proj_avg = mean(up_proj_weight, dim=0)  # [De, Dh], fixed during inference
    intermediate_avg = matmul(input_hidden, up_proj_avg.T)  # [Nt, De]
    intermediate_state -= intermediate_avg.unsqueeze(1)

    # intra-expert RMS normalization: conducted at the "dim_expert" dimension
    intermediate_state = rms_norm(intermediate_state)
    return SiLU(intermediate_state)  # standard SiLU activation
\end{lstlisting}
\end{algorithm*}

\subsection{Expert Design} \label{sec:expert-design}

\textbf{Non-gated MLP experts.} While gated MLP is widely considered superior to the non-gated variant~\cite{shazeer2020glu}, we observe that non-gated experts exhibit more favorable properties within the specific context of ReLU-based routing. Concretely, in a ReLU-activated MoE, non-gated experts obtain a more stable trend of activation ratio, whereas gated variants exhibit a sharply increasing trend. This inherent stability implies that a significantly lower regularization penalty is required to achieve a target sparsity threshold, thereby alleviating the negative impact on performance.
% Section~\ref{sec:expert-gating} provides our observations in detail.

\textbf{NormSiLU.} We introduce NormSiLU (Algorithm~\ref{alg:norm_silu}) as an enhanced activation for MoE experts, prepending a dual-stage normalization to the SiLU non-linearity within each expert.

First, inter-expert mean normalization is mathematically equivalent to centering the per-expert up-projection weights around the cross-expert mean, ensuring the pre-activation input distribution is approximately zero-centered. This adjustment stabilizes the SiLU activation distribution within experts. Second, intra-expert RMS normalization is applied to maintain consistent activation magnitudes. We find that this dual-stage normalization not only prevents internal expert activations from vanishing, but also promotes a steady activation ratio at the router level. A theoretical justification for its rationality is presented in Appendix~\ref{sec:theoretical-normsilu}.

Given the expert intermediate dimension $d_e$, the structure of a DECO expert is formally defined as:
\begin{equation}
\begin{aligned}
\label{eq:expert-design}
    \mathbf{x}_{up}&=\mathrm{SparseLinear}(\mathbf{x},\mathbf{W}_{up}),\\
    \mathbf{x}'_{up}&=\mathrm{NormSiLU}(\mathbf{x},\mathbf{W}_{up},\mathbf{x}_{up}),\\
    \mathbf{y}&=\mathrm{SparseLinear}(\mathbf{x}'_{up},\mathbf{W}_{down}),
\end{aligned}
\end{equation}
where $\mathbf{W}_{up}\in\mathbb{R}^{N_e\times d_e\times d_h}$ and $\mathbf{W}_{down}\in\mathbb{R}^{N_e\times d_h\times d_e}$ are the up-projection and down-projection weights, respectively. $\mathrm{SparseLinear}$ operator facilitates sparse linear operations by involving only active experts at inference time.

\subsection{Adaptive Sparsity Regularization} \label{sec:adaptive-reg}

To effectively control the sparsity level, we adopt an adaptive sparsity regularization, based on the router entropy loss and a dynamic scaling algorithm for the coefficient.

Router entropy is a sparsification loss applied to a normalized router score, which is calculated as:
\begin{equation}
\label{eq:router-entropy}
    \mathbf{p_1}=|\mathbf{p}|/\mathrm{Sum}(|\mathbf{p}|),
    \mathcal{L}_{ent}=-\mathbf{p}_\mathbf{1}^T\ln(\mathbf{p_1}+\epsilon)
\end{equation}
where $\mathbf{p}$ is the router score defined in Equation~\ref{eq:router-design}. The router entropy loss, $\mathcal{L}_{ent}$, is multiplied by a coefficient $\lambda$ and added to the training objective.

Instead of using a static coefficient, inspired by ReMoE~\cite{wang2024remoe}, we adaptively scale $\lambda$ according to the current sparsity level. Specifically, if the current sparsity falls below the target sparsity, $\lambda$ is scaled by a factor $\eta > 1$ for the subsequent iteration; otherwise, $\lambda$ is divided by $\eta$. In this way, DECO maintains a stable activation ratio centered precisely at the desired sparsity level.

\section{Experiments} \label{sec:experiments}

\begin{figure*}[t]
    \centering
    \includegraphics[width=\linewidth]{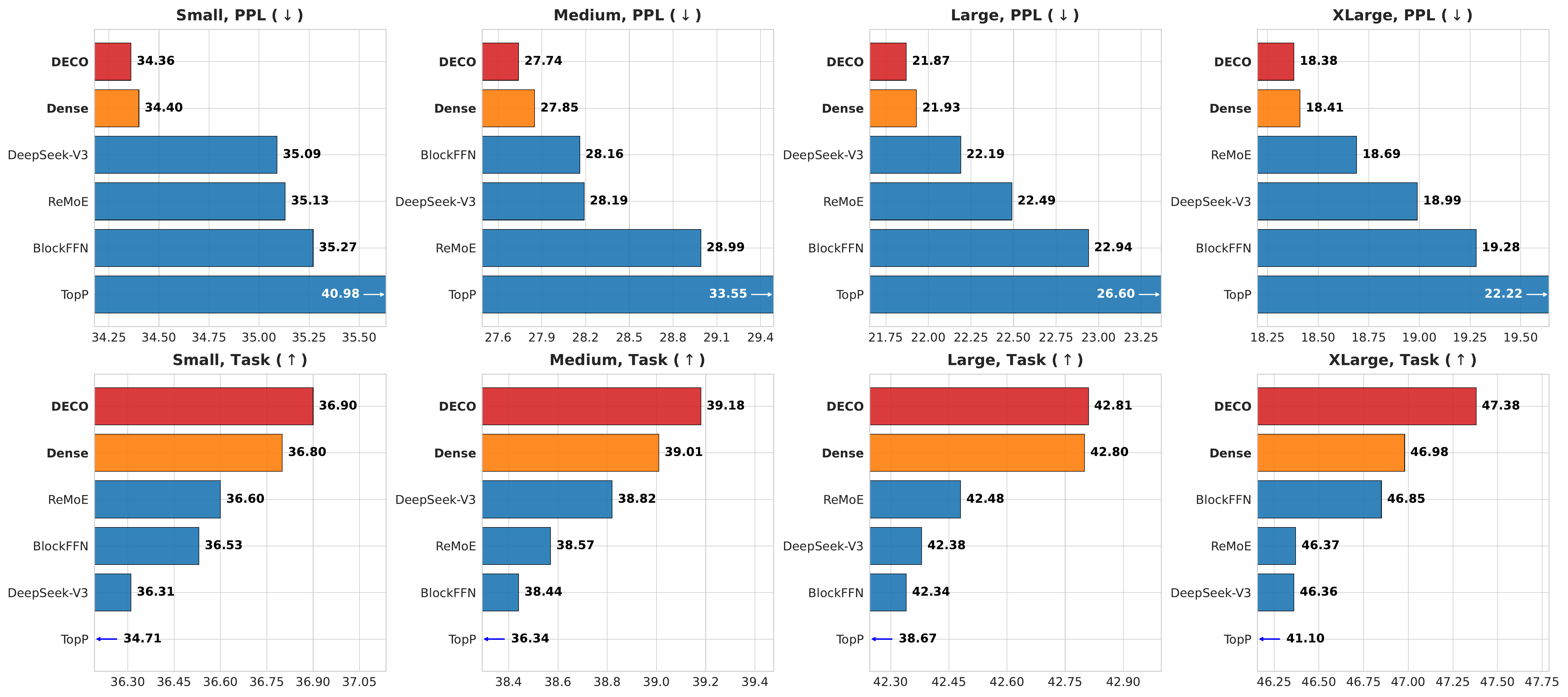}
    \caption{The evaluation results of DECO versus baseline settings. ``PPL'' and ``Task'' indicate the C4 validation perplexity and the average accuracy (\%) on downstream benchmarks, respectively. DeepSeek-V3 uses gated MLP experts, and ReMoE uses non-gated ones. This is due to their better performance than the opposite settings; see Section~\ref{sec:expert-gating} for detailed discussions.}
    \label{fig:main-result}
    \vspace{-1em}
\end{figure*}

\subsection{Main Results}

To demonstrate the architectural rationality of DECO, we introduce the following baselines: Dense (i.e., a standard LLaMA-style Transformer using SwiGLU FFNs~\cite{touvron2023llama2}), TopP~\cite{huang2024harder}, DeepSeek-V3~\cite{liu2024deepseekv3}, ReMoE~\cite{wang2024remoe}, and BlockFFN~\cite{song2025blockffn}. Four total parameter scales are involved: Small (0.11B), Medium (0.24B), Large (0.53B), and XLarge (1.18B).

All settings are trained on the same high-quality data mixture. The model performance is evaluated by two metrics: Perplexity (PPL) on the C4 English validation set~\cite{raffel2020exploring}, and average accuracy across a suite of commonsense reasoning benchmarks. See Appendix~\ref{sec:experiment-setting} for more details about experimental settings and Appendix~\ref{sec:repeated-experiments} for repeated experiments.

To ensure a rigorous comparison, within each group of \textit{the same total parameter count}, \textit{the number of training tokens is also held consistent} at around 40 times the parameter count. All non-FFN components, including attention layers and embedding layers, remain identical within each group. All MoE settings within a group share the same routed-expert activation ratio (\textit{around 20\% on the training data}) and intermediate dimension of the shared expert. Furthermore, we ensure that all routed experts have close parameter counts to maintain consistent expert granularities.

From the results shown in Figure~\ref{fig:main-result} (evaluation results on individual benchmarks are shown in Appendix~\ref{sec:independent-benchmark}), we derive the following conclusions:

(1) \textit{Dense comparability}: With an average routed-expert activation ratio of 20\%, DECO achieves performance parity with the Dense baseline. This holds true under the same total parameter budget and training token volume, demonstrating DECO's efficiency in maintaining dense-level representation power with reduced active computation.

(2) \textit{Performance superiority}: Under the same routed-expert activation ratio, shared-expert dimensions, and expert granularity, DECO surpasses existing MoE baselines in both perplexity and downstream task performance.

\begin{figure*}[ht]
\begin{minipage}{0.32\textwidth}
    \centering
    \includegraphics[width=1.0\linewidth]{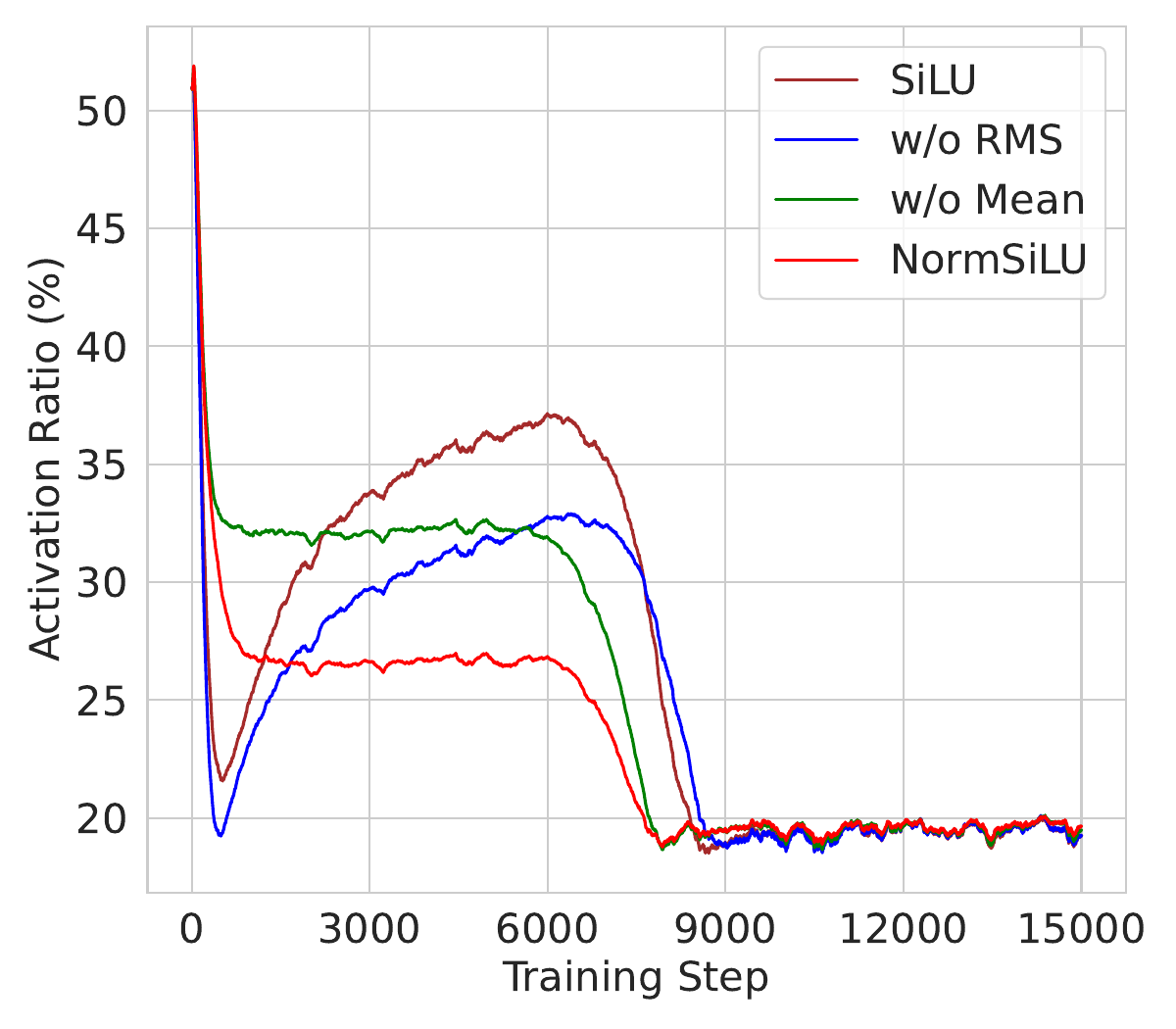}
    \caption{The trend of activation ratio of DECO (Small) and ablation settings each removing one step of NormSiLU. The baseline ``\textcolor{Brown}{SiLU}'' and ``\textcolor{Blue}{w/o RMS}'' settings show \textit{surging routed-expert activation ratio} before being pulled back by regularization.}
    \label{fig:normsilu-act}
\end{minipage}
\hfill
\begin{minipage}{0.32\textwidth}
    \centering
    \includegraphics[width=1.0\linewidth]{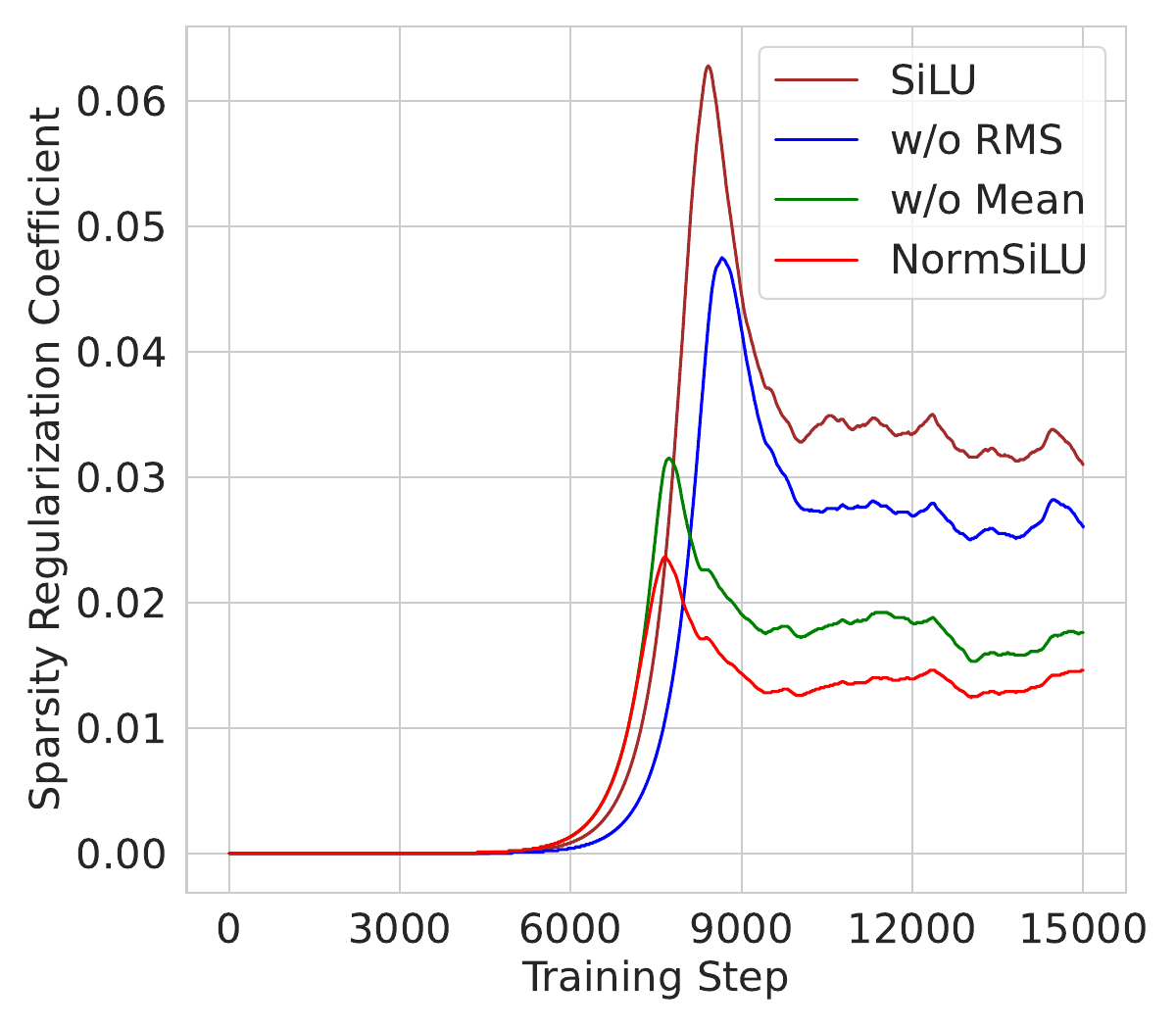}
    \caption{The trend of the regularization coefficient of DECO (Small) and ablation settings each removing one step of NormSiLU. The settings ``\textcolor{Brown}{SiLU}'' and ``\textcolor{Blue}{w/o RMS}'' show significantly higher coefficients, which potentially harm performance.}
    \label{fig:normsilu-coef}
\end{minipage}
\hfill
\begin{minipage}{0.32\textwidth}
    \centering
    \includegraphics[width=1.0\linewidth]{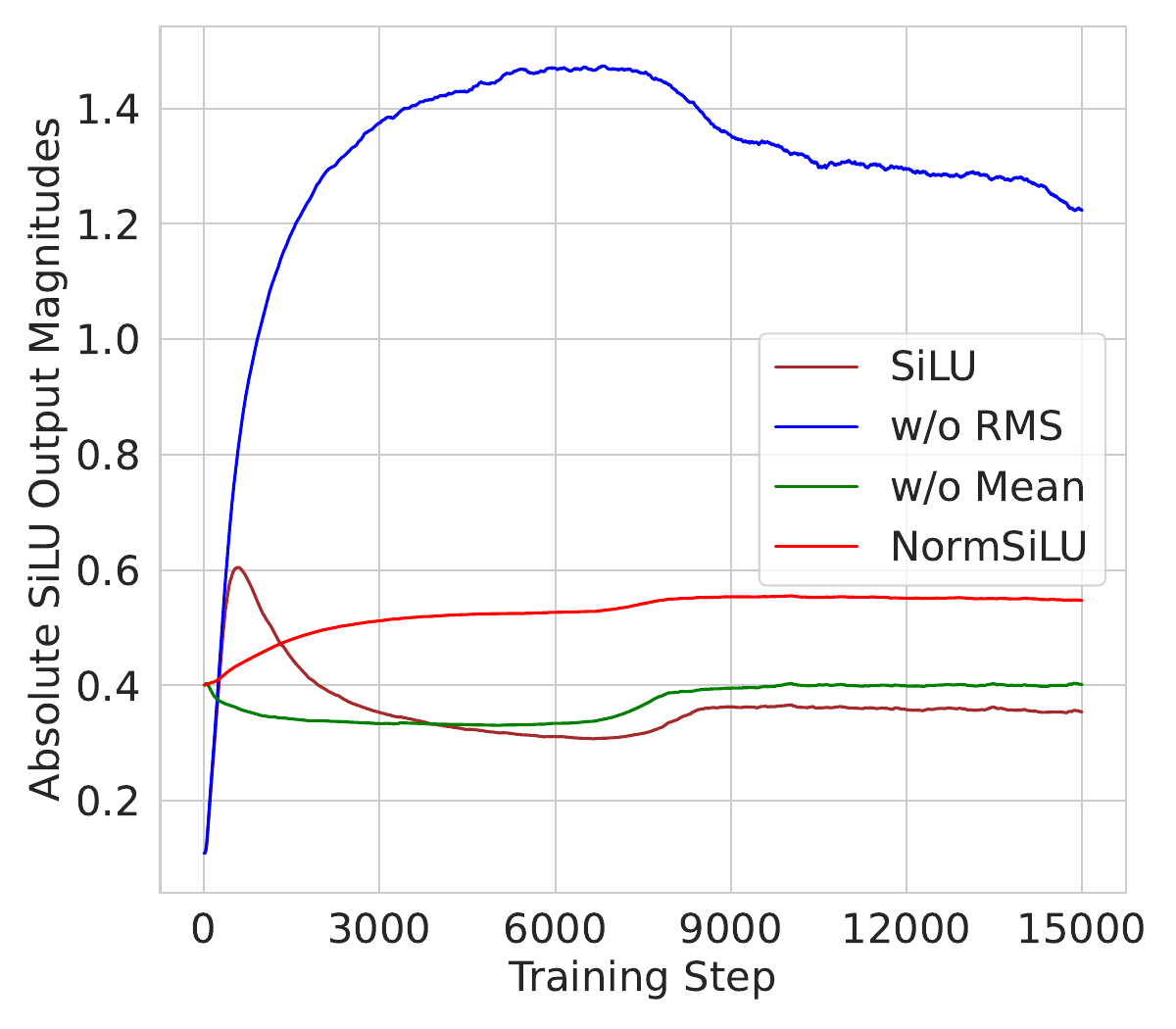}
    \caption{The average absolute output magnitudes of SiLU within routed experts of DECO (Small) and ablation settings each removing one step of NormSiLU. The settings ``\textcolor{Brown}{SiLU}'' and ``\textcolor{Green}{w/o Mean}'' show \textit{vanishing SiLU output magnitudes}.}
    \label{fig:normsilu-magnitude}
\end{minipage}
\vspace{-1em}
\end{figure*}

% \subsection{Effect of Learnable Expert-Wise Router Scaling}
\subsection{Effect of Expert-Wise Router Scaling}

\begin{table}[ht]
    \footnotesize
    \centering
    \begin{tabular}{l|cc|cc}
    \toprule
    & \multicolumn{2}{c|}{Small} & \multicolumn{2}{c}{Medium} \\
    & PPL ($\downarrow$) & Task ($\uparrow$) & PPL ($\downarrow$) & Task ($\uparrow$) \\
    \midrule
    Fixed & 34.43 & 36.74 & 27.94 & 39.08 \\
    Scalar & 35.16 & 36.63 & 27.92 & 38.54 \\
    \midrule
    \textbf{DECO} & \textbf{34.36} & \textbf{36.90} & \textbf{27.74} & \textbf{39.18} \\
    \bottomrule
    \end{tabular}
    \caption{The ablation results on the effect of learnable expert-wise router scaling. ``Fixed'' indicates one fixed scaling factor, while ``Scalar'' includes one shared learnable scalar scaling factor.}
    \label{tab:learnable-scaling}
    \vspace{-1em}
\end{table}

% \begin{figure}[ht]
%     \centering
%     \includegraphics[width=\linewidth]{figures/expert_output_scale_layer01.pdf}
%     \caption{The distribution of routed-expert output norms in the first MoE layer of DECO (Medium) on the C4 validation set, which shows expert-wise heterogeneity.}
%     \label{fig:expert-output-scale}
%     \vspace{-1em}
% \end{figure}

To demonstrate the effect of DECO's router scaling design, we experiment on two ablation settings: ``\textit{Fixed}'' adopts a constant scaling factor for all routed experts, and ``\textit{Scalar}'' involves a single learnable scalar scaling factor shared by experts. Both ablation settings are initialized with the same values as DECO. Evaluation results in Table~\ref{tab:learnable-scaling} reveal the performance benefits of learnable vectorized scaling factors. We further explore the impact of initialization value in Appendix~\ref{sec:router-norm-init}.

As empirical justification for this design, we analyze the distribution of expert output norms on the C4 validation set and find clear heterogeneity across the output scales of routed experts. For example, in the first MoE layer of DECO (Medium), the average output norm of all experts is 0.23, while the maximum value exceeds 1.14. This validates our hypothesis that applying expert-specific, learnable vectorized factors is essential to accommodate the varying output scale across different experts.

% As empirical justification for this design, we analyze the distribution of expert output norms on the C4 validation set. As illustrated in Figure~\ref{fig:expert-output-scale}, the output scales of routed experts in DECO (Medium) exhibit clear heterogeneity. These results validate our hypothesis that applying expert-specific, learnable vectorized factors is essential to accommodate the varying output scale across different experts.

\subsection{Effect of NormSiLU} \label{sec:effect-normsilu}

\begin{table}[ht]
    \footnotesize
    \centering
    \begin{tabular}{l|cc|cc}
    \toprule
    & \multicolumn{2}{c|}{Small} & \multicolumn{2}{c}{Medium} \\
    & PPL ($\downarrow$) & Task ($\uparrow$) & PPL ($\downarrow$) & Task ($\uparrow$) \\
    \midrule
    w/o Mean & 34.57 & 36.85 & 27.77 & 38.88 \\
    w/o RMS  & 35.77 & 36.78 & 27.97 & 39.04\\
    SiLU     & 35.69 & 36.46 & 28.05 & 38.85 \\
    \midrule
    NormSiLU & \textbf{34.36} & \textbf{36.90} & \textbf{27.74} & \textbf{39.18} \\
    \bottomrule
    \end{tabular}
    \caption{The ablation results on the effect of NormSiLU. The marker ``w/o'' means removing a specific normalization step, and ``SiLU'' is the vanilla SiLU operator without any normalization.}
    \label{tab:norm-silu}
    \vspace{-0.5em}
\end{table}

To validate the effect of NormSiLU, which incorporates both inter-expert mean normalization and intra-expert RMS normalization, we evaluate three ablation settings: ``\textit{w/o Mean}'' removes inter-expert mean normalization, ``\textit{w/o RMS}'' removes intra-expert RMS normalization, and ``\textit{SiLU}'' is the standard SiLU operator without any normalization.

As shown in Table~\ref{tab:norm-silu}, both normalization steps contribute positively, with intra-expert RMS normalization providing a more substantial gain. To investigate the underlying mechanisms of NormSiLU, we track three critical variables throughout the training process: the routed-expert activation ratio, the sparsity regularization coefficient, and the average absolute output magnitudes of SiLU within experts.

As illustrated in Figure~\ref{fig:normsilu-act}, the activation ratios of ``SiLU'' and ``w/o RMS'' surge rapidly during the initial training phase. While the adaptive regularization pulls back this surge, Figure~\ref{fig:normsilu-coef} reveals that these settings require a significantly higher regularization coefficient, which typically degrades overall performance. Conversely, NormSiLU and ``w/o Mean'' maintain stable activation trends, with NormSiLU achieving the lowest activation ratio. We conclude that \textbf{intra-expert RMS normalization mitigates the uncontrolled growth of activation ratios, while inter-expert mean normalization further promotes sparsity level}.

Moreover, analysis of the internal SiLU output magnitudes (Figure~\ref{fig:normsilu-magnitude}) reveals that ``SiLU'' and ``w/o Mean'' exhibit considerably lower magnitudes. This suggests that expert neurons (i.e., parameter columns/rows) in these settings are potentially under-utilized and less significantly activated. In contrast, \textbf{inter-expert mean normalization effectively addresses this issue, ensuring more robust activation and utilization of expert neurons}.

% On the other hand, considering the output magnitudes of SiLU operators inside experts, Figure~\ref{fig:normsilu-magnitude} shows that ``SiLU'' and ``w/o Mean'' settings have considerably low SiLU output magnitudes. This indicates that the neurons (i.e., parameter columns/rows) of experts in these two settings are less significantly activated and utilized. \textbf{Inter-expert mean normalization can effectively resolve this problem, making SiLU activate expert neurons more significantly.}

\subsection{Effect of Expert Gating} \label{sec:expert-gating}

\begin{table}[ht]
    \footnotesize
    \centering
    \setlength{\tabcolsep}{0.2em}
    \begin{tabular}{c|cc|cc|cc}
    \toprule
    \multirow{2}{*}{Setting} & \multicolumn{2}{c|}{Small} & \multicolumn{2}{c|}{Medium} & \multicolumn{2}{c}{Large} \\
    & PPL ($\downarrow$) & Task & PPL ($\downarrow$) & Task & PPL ($\downarrow$) & Task \\
    \midrule
    DS-V3 (GA) & \textbf{35.09} & 36.31 & \textbf{28.19} & \textbf{38.82} & \textbf{22.19} & \textbf{42.38} \\
    DS-V3 (NG) & 35.16 & \textbf{36.73} & 28.28 & 38.47 & 22.33 & 42.02 \\
    \midrule
    ReMoE (GA) & 36.02 & \textbf{36.61} & 29.55 & 38.15 & 22.78 & 42.27 \\
    ReMoE (NG) & \textbf{35.13} & 36.60 & \textbf{28.99} & \textbf{38.57} & \textbf{22.49} & \textbf{42.48} \\
    \midrule
    DECO (GA) & 39.77 & 35.78 & 32.46 & 37.29 & 22.13 & 42.29 \\
    DECO (NG) & \textbf{34.36} & \textbf{36.90} & \textbf{27.74} & \textbf{39.18} & \textbf{21.87} & \textbf{42.81} \\
    \bottomrule
    \end{tabular}
    \caption{The ablation results on the effect of expert gating. ``GA'' and ``NG'' indicate gated MLP experts and non-gated MLP experts, respectively. ``DS-V3'' indicates DeepSeek-V3.}
    \label{tab:expert-gating}
    \vspace{-1em}
\end{table}

\begin{figure}[ht]
    \centering
    \includegraphics[width=0.9\linewidth]{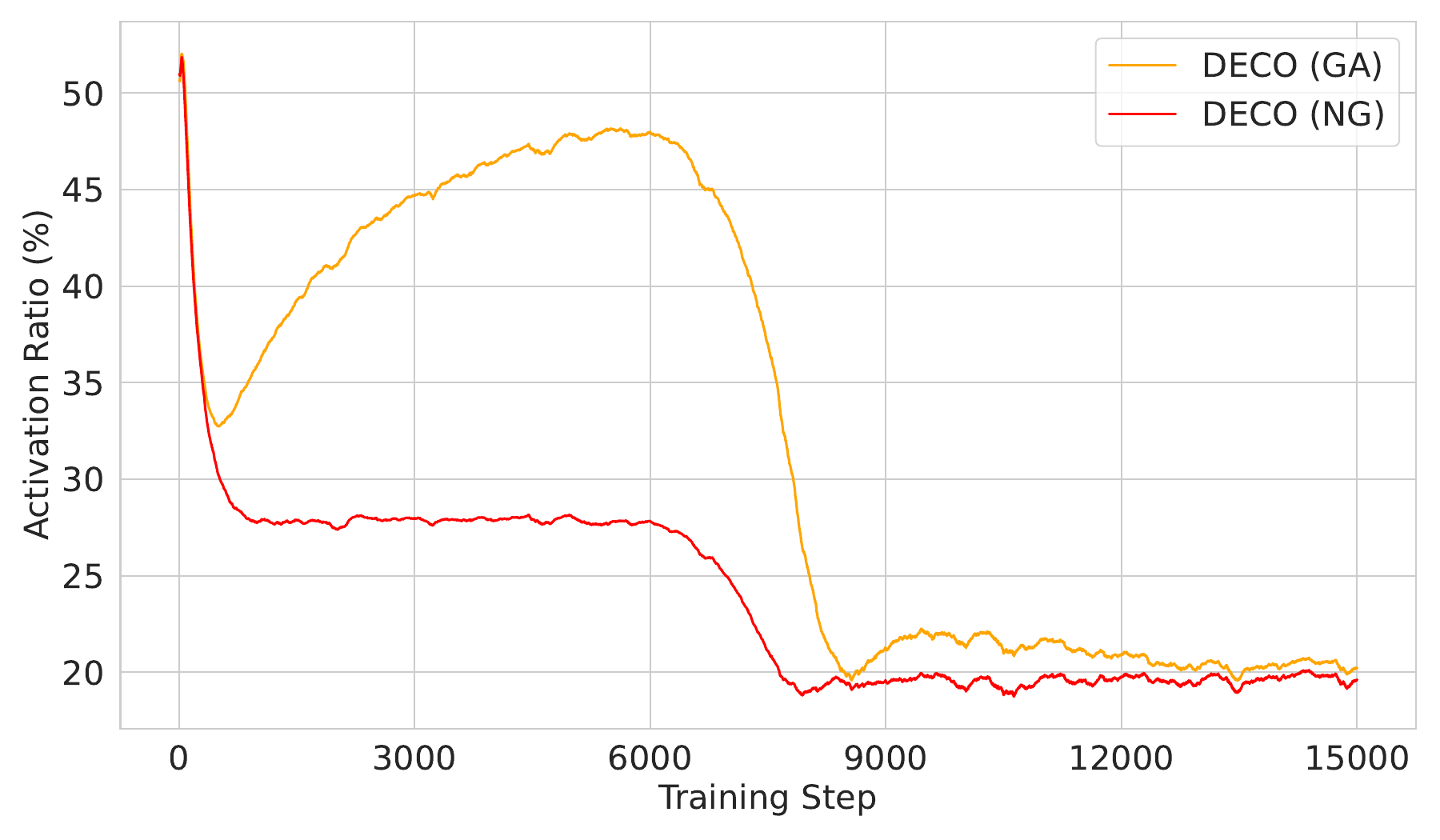}
    \caption{The trend of routed-expert activation ratio of DECO (Small) using different expert gating policies.}
    \label{fig:expert-gating-act}
    \vspace{-1em}
\end{figure}

To investigate whether expert gating significantly influences MoE performance, we conduct experiments on three architectures: DeepSeek-V3, ReMoE, and DECO. DeepSeek-V3 is a well-performing MoE architecture using TopK routing, while ReMoE and DECO use ReLU-based routing to implement a flexible activation ratio. For each architecture, we compare non-gated MLP experts (NG) against gated MLP experts (GA).

As demonstrated by Table~\ref{tab:expert-gating}, \textbf{for ReLU-based routing, non-gated MLP experts generally surpass gated counterparts}. Conversely, for standard TopK routing (e.g., DeepSeek-V3), gated experts provide a marginal performance gain, though the difference is not significant.

We attribute this disparity to the training dynamics that emerge when gated experts are paired with flexible, threshold-based routing. As illustrated in Figure~\ref{fig:expert-gating-act}, DECO (GA) exhibits highly unstable activation trends, characterized by a drastic surge in the routed-expert activation ratio that must be aggressively counteracted by sparsity regularization. In contrast, DECO (NG) maintains a stable activation trajectory, requiring substantially less regularization and thereby preserving model performance.

Mechanistically, this divergence may stem from gradient behavior. Compared with non-gated variants, gated experts (e.g., SwiGLU) contain more multiplicative interactions that produce highly dynamic output scales, sending massive gradient signals back to the router. Because ReLU-based routing couples activation directly to the logit threshold, this gradient surge drastically destabilizes the activation ratio. Conversely, in TopK-routed architectures like DeepSeek-V3, the hard constraint of activating a fixed number of experts per token effectively masks this logit-induced instability, rendering the overall performance largely insensitive to the choice of expert gating.

\subsection{Effect of Activation Ratio} \label{sec:effect-activation-ratio}

\begin{figure}
    \centering
    \includegraphics[width=0.9\linewidth]{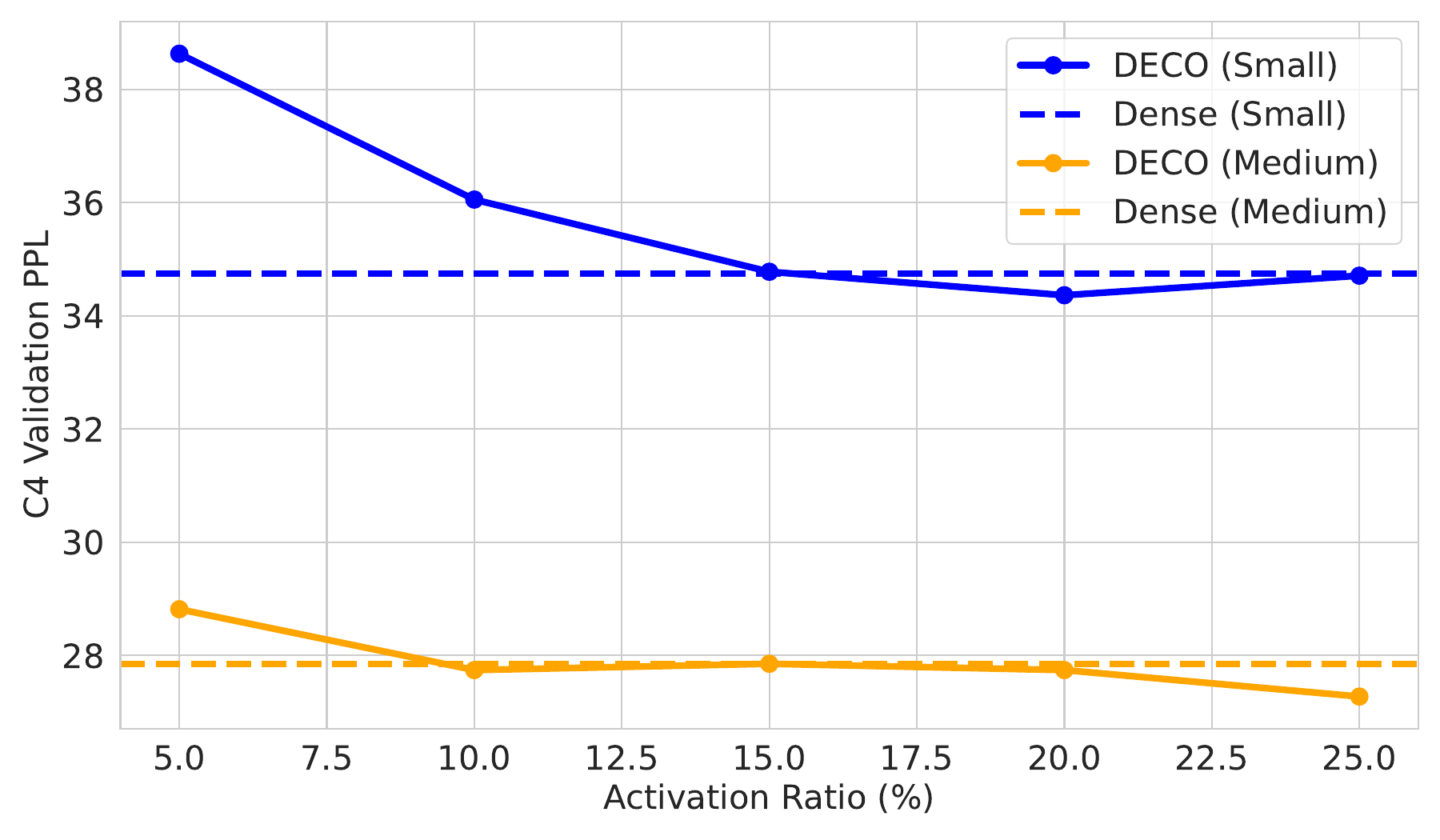}
    \caption{The impact of the routed-expert activation ratio on the performance of DECO.}
    \label{fig:effect-act-ratio}
    \vspace{-1em}
\end{figure}

We evaluate DECO across a wide range of routed-expert activation ratios, from 5\% to 25\%. The experimental results, illustrated in Figure~\ref{fig:effect-act-ratio}, yield two primary observations:

% Notably, according to the red line in Figure~\ref{fig:expert-gating-act}, DECO exhibits an intrinsic activation ratio of approximately 28\% if no regularization is applied. Therefore, reaching target ratios beyond this intrinsic threshold is infeasible.

(1) Positive correlation with performance: DECO's performance scales positively with the activation ratio. This aligns with findings in standard TopK MoE, where increased activation ratios correspond to higher per-token computational investment, typically resulting in superior performance.

(2) Scale-dependent comparability thresholds: The activation ratio required for DECO to achieve parity with dense models varies by scale. Specifically, the ``Small'' setting reaches dense-level performance at a 15\% activation ratio, whereas ``Medium'' requires only 10\%. This suggests that as DECO scales up, it may attain dense-comparable parameter efficiency at lower activation ratios. This observation is consistent with recent literature suggesting that the optimal activation ratio decreases as total parameter count increases~\cite{zhao2025towards}. Further investigation on larger-scale models is required to validate this scaling trend.

Besides activation ratio, we also study the effect of other MoE-specific factors, including expert granularity and shared expert size; see Appendix~\ref{sec:key-factor}.

\begin{table*}[t]
    \footnotesize
    \centering
    \begin{tabular}{c|c|cccccc|c}
    \toprule
    Device & Setting & \textbf{MT.} & \textbf{Trans.} & \textbf{Summ.} & \textbf{QA} & \textbf{Math} & \textbf{RAG} & \textbf{Average} \\
    \midrule
    & Dense & 86.79 & 88.01 & 86.38 & 88.08 & 87.98 & 86.42 & \textbf{87.28} \\
    RTX 4090 & TopK & 194.12 & 203.75 & 193.27 & 196.70 & 195.97 & 198.12 & \textbf{196.99} \\
    24GB & DECO & 223.38 & 221.76 & 217.34 & 227.56 & 230.94 & 221.49 & \textbf{223.75} \\
    & Speedup & 2.57$\times$ & 2.52$\times$ & 2.52$\times$ & 2.58$\times$ & 2.62$\times$ & 2.56$\times$ & \textbf{2.56$\times$} \\
    \midrule
    & Dense & 15.00 & 15.21 & 14.93 & 15.24 & 15.22 & 14.94 & \textbf{15.09} \\
    Jetson AGX Orin & TopK & 39.28 & 41.70 & 38.94 & 40.03 & 40.12 & 40.16 & \textbf{40.04} \\
    64GB & DECO & 44.05 & 43.99 & 42.69 & 45.21 & 45.92 & 43.57 & \textbf{44.24} \\
    & Speedup & 2.94$\times$ & 2.89$\times$ & 2.86$\times$ & 2.97$\times$ & 3.02$\times$ & 2.92$\times$ & \textbf{2.93$\times$} \\
    \bottomrule
    \end{tabular}
    \caption{Single-GPU decoding speeds on Spec-Bench (token/sec) and average speedup ratios of our acceleration kernel (denoted as ``DECO'') relative to ``Dense'' on different devices.}
    \label{tab:kernel-speedup}
    \vspace{-1em}
\end{table*}

\section{Practical Inference Acceleration}

To demonstrate the practical utility of DECO for real-world deployment, we implement an acceleration kernel tailored for DECO, which leverages sparse activation to reduce both computational overhead and memory access latency incurred by inactive routed experts. We evaluate the kernel’s efficacy on Spec-Bench~\cite{xia2024unlocking}, a comprehensive acceleration benchmark, using a single NVIDIA RTX 4090 (24GB) and Jetson AGX Orin (64GB) as the inference devices. We establish two baselines: \textbf{Dense}, which employs standard autoregressive decoding without sparsity optimizations, and \textbf{TopK}, which substitutes our ReLU-based routing with conventional TopK. To ensure a strictly controlled evaluation, all baselines and our proposed acceleration kernel are implemented within FR-Spec~\cite{zhao2025fr}, a high-performance, CUDA-optimized inference framework. Both the TopK baseline and our kernel share an identical average activation ratio of 20\%.

As detailed in Table~\ref{tab:kernel-speedup}, our acceleration kernel substantially outperforms the Dense baseline, achieving an average speedup of 2.56$\times$ on an RTX 4090 and 2.93$\times$ on a Jetson AGX Orin. These results demonstrate that the sparse activation patterns induced by DECO effectively translate into tangible throughput improvements during practical inference. Furthermore, our kernel also outperforms standard TopK under identical sparsity levels. This efficiency advantage stems from the element-wise, in-place execution of the ReLU activation, which incurs a fundamentally lower computational complexity than the vector-wise sorting operations necessitated by TopK.

\section{Discussion}

\textbf{Considering the intrinsic activation sparsity of ``dense'' language models, it is reasonable to expect MoE to be dense-comparable.} The property of activation sparsity, indicating that only a small fraction of parameters contribute largely to final outputs, also exists in dense models. Recent studies~\cite{song2024prosparse,luo2024sparsing,zhang2024relu} indicate that for each token, only about 30\%$\sim$40\% of neurons within a standard SwiGLU FFN provide non-negligible contributions. The remaining 60\%$\sim$70\% of neurons ``seem'' to occupy computation, but actually contribute little and receive negligible gradient updates from the optimizer due to near-zero SiLU activation values. Actually, dense models can be regarded as a special form of sparse MoE, where the SiLU-activated gating projection functions as a ``router'', and neurons within up-down projections serve as ``experts''. Therefore, given an optimized architecture and a sufficient activation ratio, it is possible for a sparse MoE model to match dense performance.

% \begin{table}[ht]
%     \footnotesize
%     \centering
%     % \setlength{\tabcolsep}{0.3em}
%     \begin{tabular}{l|cc|cc}
%     \toprule
%     & \multicolumn{2}{c|}{Small} & \multicolumn{2}{c}{Medium} \\
%     & PPL ($\downarrow$) & Task ($\uparrow$) & PPL ($\downarrow$) & Task ($\uparrow$) \\
%     \midrule
%     Dense & \textbf{29.76} & 38.03 & 24.85 & 41.49 \\
%     DECO & 30.14 & \textbf{38.65} & \textbf{24.80} & \textbf{41.58}  \\
%     \bottomrule
%     \end{tabular}
%     \caption{The experimental results on FineWeb, a dataset with less heterogeneous composition.}
%     \label{tab:finweb-results}
%     % \vspace{-0.5em}
% \end{table}

\textbf{Dense-comparability may be easier to achieve under more heterogeneous training data.} In previous experiments, we adopt a diverse data mixture for training, including web texts, code, math, and many other data categories. On such heterogeneous data, DECO matches or exceeds the dense baselines across various parameter scales (Figure~\ref{fig:main-result}). However, this ``dense-comparability'' may be sensitive to the underlying data composition. When trained on FineWeb~\cite{penedo2024fineweb}, a less heterogeneous dataset, DECO (Small) has a slightly higher but close PPL compared with Dense (Small) (30.14 vs. 29.76), though exceeding Dense in task accuracy (38.65 vs. 38.03). A reasonable explanation is that high-entropy, multi-domain datasets are inherently better suited for sparse MoE. In such settings, the model can effectively process domain-specific tasks by activating only a specialized subset of its parameters within MoE layers. More studies are needed to verify this explanation.

% \textbf{Dense-comparability may be easier to achieve under more heterogeneous training data.} In previous experiments, we adopt a diverse data mixture for training, including web texts, code, math, and many other data categories. On such heterogeneous data, DECO matches or exceeds the dense baselines across various parameter scales (Figure~\ref{fig:main-result}). However, this ``dense-comparability'' may be sensitive to the underlying data composition. As shown in Table~\ref{tab:finweb-results}, when trained on FineWeb~\cite{penedo2024fineweb}, a less heterogeneous dataset, the dense-comparability of DECO is less significant, and DECO (Small) slightly lags behind Dense (Small) in PPL. A reasonable explanation is that high-entropy, multi-domain datasets are inherently better suited for sparse MoE. In such settings, the model can effectively process domain-specific tasks by activating only a specialized subset of its parameters. More studies are needed to verify this explanation.

\section{Conclusion}

In this work, we propose DECO, a novel MoE architecture designed to minimize computational and storage overhead while maintaining dense-comparable performance. Under an activation ratio of 20\%, DECO not only matches the performance of dense models with an equivalent parameter count and training token budget, but also consistently outperforms existing MoE baselines. Our analysis demonstrates that DECO's success stems from a series of architectural refinements, including ReLU-based routing with learnable expert-wise scaling, as well as the integration of non-gated MLP experts utilizing the NormSiLU activation. Furthermore, practical acceleration is achieved on real hardware, confirming that the sparsity of DECO can be translated into tangible efficiency gains for deployment.

\section*{Limitations}

One potential limitation of this work is that we do not conduct experiments involving the supervised fine-tuning (SFT) or reinforcement learning (RL) stages. Recent studies have indicated that MoE architectures may encounter unique challenges during post-training, such as RL instability resulting from fluctuating router activations~\cite{zheng2025group}. Therefore, it is reasonable to assume that DECO may have similar issues. To address this limitation, we are currently training a larger, product-level DECO model optimized for edge-device deployment. This process will involve identifying potential issues during the stages of SFT and RL, and developing corresponding mitigation strategies.

Moreover, as already stated in this article, several hypotheses and empirical observations deserve more extensive verification. Specifically, it remains to be determined how the activation ratio threshold required for dense-comparability scales with model size, and how DECO’s intrinsic sparsity and performance fluctuate across diverse data distributions or inference tasks. Investigating these factors is essential to further establish the robustness and theoretical foundation of our proposed architecture.

\section*{Ethics Statement}

This work presents DECO, a sparse MoE architecture optimized for end-side devices. By significantly reducing the active parameter scale and memory-access overhead without compromising representation power, our architecture promotes more energy-efficient deployment of LLMs. This efficiency translates to a reduced carbon footprint during inference and helps democratize advanced AI capabilities by making them accessible on resource-constrained edge hardware.

While our architectural improvements are fundamentally algorithmic, we also rely on large-scale web corpora (e.g., FineWeb, the Pile) and will inevitably inherit the historical biases, toxicity, and representational harms present in the pre-training data. We have conducted cleaning and detoxification to the data and used a rule-based program to remove personal information and offensive contents. We encourage researchers and practitioners utilizing our architecture to apply rigorous safety alignment and bias mitigation protocols before releasing downstream models. We use AI agents to refine our writing of this article.

% This paper presents work whose goal is to advance the field of Natural Language Processing. There are many potential social consequences of our work, none of which we feel must be specifically highlighted here.

\bibliography{custom}

\clearpage
\appendix

\section{Effect of Key MoE Hyperparameters} \label{sec:key-factor}

Compared with dense architectures, MoE models introduce unique hyperparameters that significantly influence performance, among which the activation ratio, expert granularity, and shared expert size are often considered the most important ones~\cite{tian2025towards}. Similarly, the performance of DECO is also sensitive to these factors, and the dense-comparability of DECO holds only under certain conditions. We have already studied the effect of the routed-expert activation ratio in Section~\ref{sec:effect-activation-ratio}, which has a significant impact on MoE performance. In this section, we characterize the impact of the remaining two MoE-specific factors.

% \begin{figure*}[t]
% \begin{minipage}{0.32\textwidth}
%     \centering
%     \includegraphics[width=1.0\linewidth]{figures/valid_act_curve_97.pdf}
%     \caption{The impact of the routed-expert activation ratio on the performance of DECO.}
%     \label{fig:effect-act-ratio}
% \end{minipage}
% \hfill
% \begin{minipage}{0.32\textwidth}
%     \centering
%     \includegraphics[width=1.0\linewidth]{figures/valid_share_curve_97.pdf}
%     \caption{The impact of shared expert sizes on performance. Routed expert dimension is 64.}
%     \label{fig:effect-share-expert}
% \end{minipage}
% \hfill
% \begin{minipage}{0.32\textwidth}
%     \centering
%     \includegraphics[width=1.0\linewidth]{figures/valid_granularity_curve_97.pdf}
%     \caption{The impact of the expert granularity ($g=4d_h/d_e$) on performance of DECO.}
%     \label{fig:effect-granularity}
% \end{minipage}
% \vspace{-1em}
% \end{figure*}

\subsection{Shared Expert Size}

\begin{figure}[ht]
    \centering
    \includegraphics[width=1.0\linewidth]{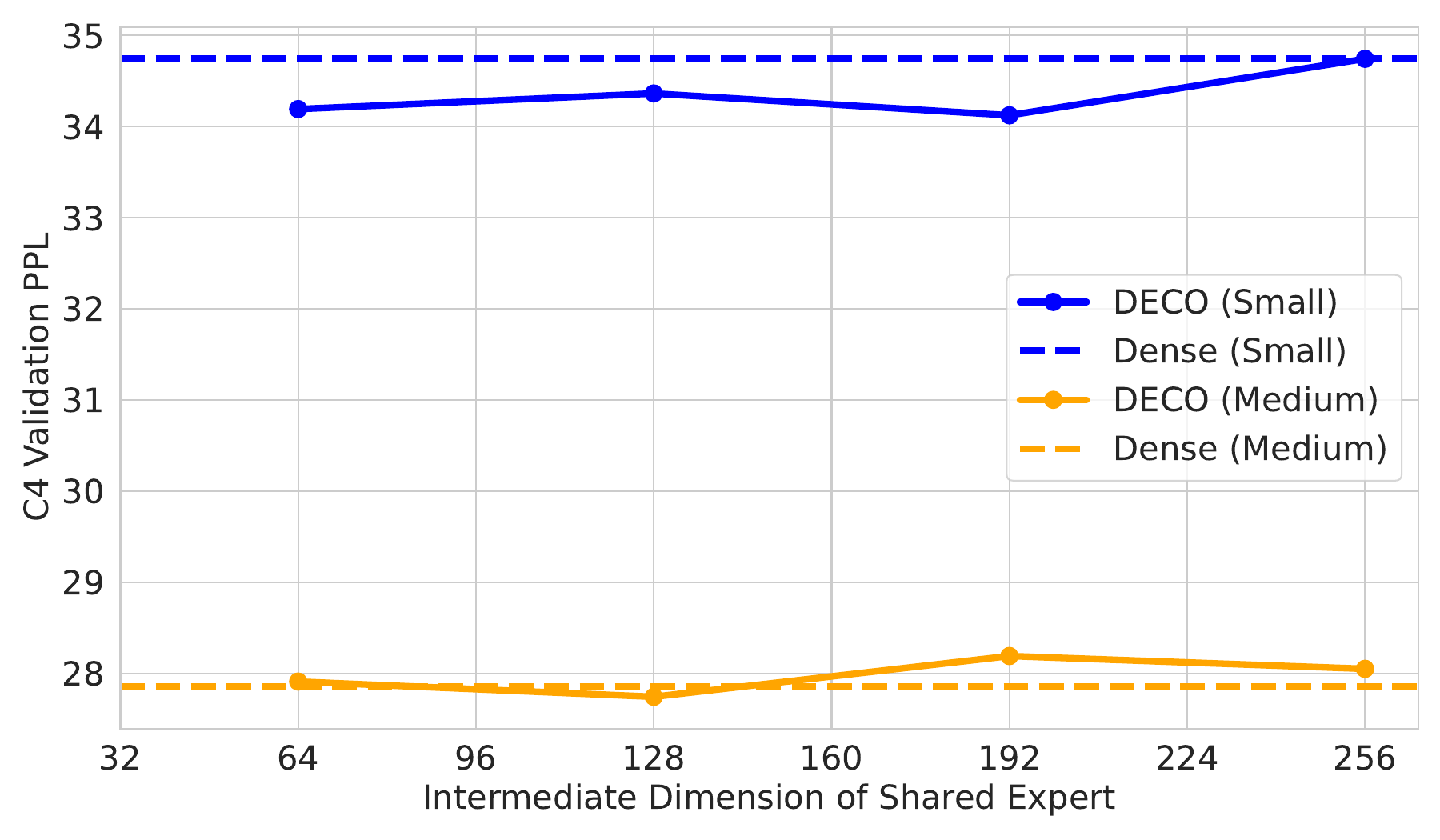}
    \caption{The impact of shared expert sizes on performance. Routed expert dimension is 64.}
    \label{fig:effect-share-expert}
\end{figure}

We evaluate the performance of DECO across various shared-expert sizes, proportional to the intermediate dimension of the shared expert. For rigorous comparison, we hold the total parameter count approximately constant. This necessitates a trade-off between the capacity of the shared expert and the number of routed experts.

As illustrated in Figure~\ref{fig:effect-share-expert}, when the intermediate dimension of the shared expert is around 1$\sim$2 times that of a routed expert (fixed at 64), DECO achieves comparability with the dense baseline and exhibits relative insensitivity to the size of the shared expert. However, as the shared-expert dimension increases to 3$\sim$4 times that of the routed experts, the perplexity (PPL) degrades significantly. This performance drop is attributed to the reduced number of routed experts necessitated by the fixed parameter budget. These findings corroborate the observations in~\citet{tian2025towards}, suggesting that an oversized shared expert is unnecessary. Instead, a single shared expert with a scale comparable to that of a routed expert appears to be optimal.

\subsection{Expert Granularity}

\begin{figure}[ht]
    \centering
    \includegraphics[width=1.0\linewidth]{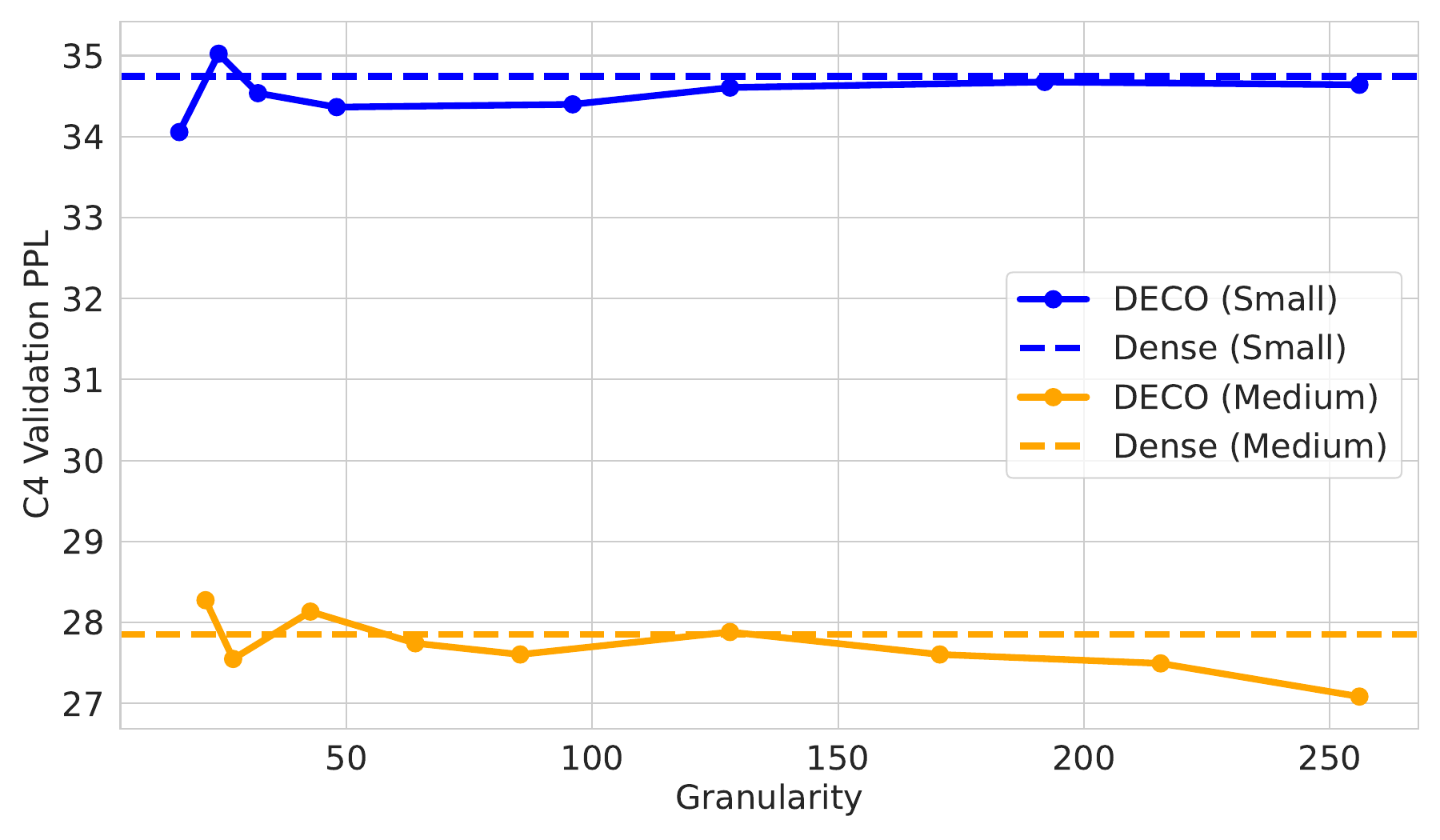}
    \caption{The impact of the expert granularity ($g=4d_h/d_e$) on performance of DECO.}
    \label{fig:effect-granularity}
\end{figure}

As illustrated in Figure~\ref{fig:effect-granularity}, for ``Medium'' setting, we observe a monotonic improvement in perplexity (PPL) as granularity increases when $g>120$. However, the ``Small'' setting shows reduced sensitivity to granularity, as it is potentially more difficult for its lower-capacity experts to capture the ``long-tail'' distribution of the data or to develop fine-grained specialization. For both settings, when granularity is below 60, performance fluctuates near the dense baseline. By contrast, for $g>60$, DECO consistently achieves dense parity. These results suggest that finer expert granularity is essential for stable and competitive performance.

\begin{table*}[ht]
    \footnotesize
    \centering
    \setlength{\tabcolsep}{0.4em}
    \begin{tabular}{l|cccccccccccc}
    \toprule
    Scale & $d_h$ & $d_e$ & $d_s$ & $N_e$ & $d_{ff}$ & $N_{layer}$ & $n_{step}$ & $lr$ & $batch\_size$ & $n_{tokens}$ & $\eta$ & $\lambda_{init}$ \\
    \midrule
    0.11B (Small) & 768 & 64 & 128 & 42 & 1,920 & 16 & 15,000 & $1.175e-3$ & $2.62\times 10^5$ & $3.93\times 10^9$ & 1.002 & $1e-8$ \\
    0.24B (Medium) & 1,024 & 64 & 128 & 57 & 2,560 & 20 & 15,000 & $9.3e-4$ & $5.24\times 10^5$ & $7.86\times 10^9$ & 1.002 & $1e-8$ \\
    0.53B (Large) & 1,280 & 64 & 128 & 77 & 3,360 & 27 & 20,000 & $7.8e-4$ & $1.05\times 10^6$ & $2.10\times 10^{10}$ & 1.001 & $1e-8$ \\
    1.18B (XLarge) & 1,792 & 64 & 128 & 102 & 4,480 & 32 & 23,000 & $6.54e-4$ & $2.10\times 10^6$ & $4.83\times 10^{10}$ & 1.001 & $1e-8$ \\
    \bottomrule
    \end{tabular}
    \caption{The hyperparameter settings of models. $d_s$, $d_{ff}$, $N_{layer}$, $n_{step}$, and $n_{tokens}$ denote the intermediate dimension of the shared expert, the intermediate dimension of the dense FFN layer, the total number of hidden layers, the number of training steps, and the number of training tokens, respectively. $\eta$ and $\lambda_{init}$ are the coefficient multiplier and the initial value of the regularization coefficient as used in the adaptive sparsity regularization.}
    \label{tab:model-setting}
    \vspace{-0.5em}
\end{table*}

\begin{table*}[ht]
    \footnotesize
    \centering
    \setlength{\tabcolsep}{0.5em}
    \begin{tabular}{l|cccc}
    \toprule
    Setting & 5-run PPL ($\downarrow$) & Average PPL ($\downarrow$) & 5-run Task ($\uparrow$) & Average Task ($\uparrow$) \\
    \midrule
    Dense (Small) & 34.40 34.28 34.49 34.59 34.38 & \textbf{34.43$\pm$0.105} & 36.80 36.93 36.71 36.67 36.81 & 36.78$\pm$0.090 \\
    DECO (Small)  & 34.36 34.51 34.90 34.91 34.75 & 34.69$\pm$0.218 & 36.90 36.78 36.73 37.15 36.65 & \textbf{36.84$\pm$0.174} \\
    \midrule
    Dense (Medium) & 27.85 27.77 27.15 27.79 28.07 & 27.73$\pm$0.307 & 39.01 38.85 39.14 38.96 38.93 & 38.98$\pm$0.096 \\
    DECO (Medium)  & 27.74 27.17 27.71 27.22 27.81 & \textbf{27.53$\pm$0.276} & 39.18 39.35 38.91 39.37 38.85 & \textbf{39.13$\pm$0.217} \\
    \bottomrule
    \end{tabular}
    \caption{The repeated experimental results on the ``Small'' and ``Medium'' scales.}
    \label{tab:repeated-results}
    \vspace{-1em}
\end{table*}

\section{Detailed Experimental Settings} \label{sec:experiment-setting}

\paragraph{Training datasets} For all settings, we use the same pretraining dataset, which consists of a mixture of various data, including FineWeb~\cite{penedo2024fineweb}, Nemotron-CC~\cite{su2025nemotron}, Pile~\citep{gao2020pile}, Wikipedia, and many other collected raw corpora. The training data covers a wide range of categories, such as raw web text, math, and code. The data mixing ratio is carefully tuned through experiments on small-scale models.

\paragraph{Evaluation benchmarks} We evaluate the accuracy of models on various commonsense reasoning benchmarks with LMEval~\cite{eval-harness}, including PIQA~\citep{piqa}, SIQA~\citep{siqa}, HellaSwag~\citep{hellaswag}, ARC-C, ARC-E~\cite{arc}, WinoGrande~\cite{winogrande}, and LAMBADA~\cite{lambada}.

\paragraph{Baseline adjustments} Since BlockFFN is originally designed to combine activation sparsity and speculative decoding for better acceleration, it introduces the chunk sparsification loss and activation locality loss to increase the union sparsity level of consecutive tokens as well as the activation similarity between neighboring tokens~\cite{song2025blockffn}. However, as we do not target these objectives in this work, for fair comparison, we replace its original regularization with our own adaptive sparsity regularization.

\begin{table}[ht]
    \footnotesize
    \centering
    \begin{tabular}{l|cc}
    \toprule
    Setting & PPL ($\downarrow$) & Task ($\uparrow$) \\
    \midrule
    Small (two-loss)  & 35.62 & 36.23 \\
    Small (one-loss)  & \textbf{35.27} & \textbf{36.53} \\
    Medium (two-loss) & 28.65 & 38.18 \\
    Medium (one-loss) & \textbf{28.16} & \textbf{38.44} \\
    \bottomrule
    \end{tabular}
    \caption{The performance of BlockFFN using its original regularization (two-loss) and our own adaptive sparsity regularization (one-loss).}
    \label{tab:blockffn-adjust}
    \vspace{-1.5em}
\end{table}

To justify such an adjustment, we implement BlockFFN strictly using its original ``two-loss'' setting, tuning the loss coefficients via grid search. From the results in Table~\ref{tab:blockffn-adjust}, at the same target activation ratio of 20\%, our ``one-loss'' setting is consistently better than the original ``two-loss'' regularization, demonstrating the rationality of our experimental setting for BlockFFN.

\paragraph{Hyperparameters} To find the optimal hyperparameters, we first conduct a grid search on small-scale models. Then, following DeepSeek LLM~\cite{bi2024deepseek}, we assume a power-law relationship of both the optimal learning rate and batch size with respect to the computation budget. Thereby, we can extrapolate the hyperparameters for large-scale models using those for small-scale models. The detailed experimental hyperparameters are shown in Table~\ref{tab:model-setting}. All MoE baseline settings and DECO use a dense FFN at the first layer, and adopt sparse MoE for the remaining $N_{layer}-1$ layers. We use the WSD learning rate scheduler along the training process~\cite{hu2024minicpm,dubey2024llama}, with 100 warmup steps at the beginning and 1,000 decay steps at the last stage.

\section{Impact of the Scaling Factor Initialization Value} \label{sec:router-norm-init}

\begin{table}[ht]
    \footnotesize
    \centering
    \setlength{\tabcolsep}{0.3em}
    \begin{tabular}{l|ccccccc}
    \toprule
    Init Value & 0.01 & 0.025 & 0.05 & 0.1 & 0.25 & 0.5 & 1.0 \\
    \midrule
    PPL ($\downarrow$) & 34.71 & 35.66 & 34.61 & 34.36 & \textbf{34.26} & 35.20 & 34.74 \\
    Task ($\uparrow$)  & 36.86 & 36.57 & 36.64 & \textbf{36.90} & 36.72 & 36.57 & 36.54 \\
    \bottomrule
    \end{tabular}
    \caption{The performance of DECO (Small) when the learnable expert-wise scaling factors are initialized with different values.}
    \label{tab:router-norm-init}
    \vspace{-1em}
\end{table}

In this section, we conduct experiments on DECO when its learnable expert-wise scaling factors are initialized with different values. The results are shown in Table~\ref{tab:router-norm-init}. The performance of DECO does not change monotonically. Empirically, the best performance is achieved when the scaling factors are initialized around 0.1$\sim$0.25. In our experiments in Section~\ref{sec:experiments}, we choose the initialization value of 0.1 for all DECO settings.

\section{Repeated Experiments} \label{sec:repeated-experiments}

% Due to the expensive GPU costs required by large-scale pre-training, most pre-training experiments are run once. To further promote the reliability of our results, we conduct repeated experiments on the scales of ``Small'' and ``Medium''. Each scale is run five times with different random seeds. From the results in Table~\ref{tab:repeated-results}, DECO presents stably comparable performance with Dense.

Due to the prohibitive GPU costs associated with large-scale pre-training, experiments in this domain are typically executed only once. To rigorously validate the reliability of our findings, we conducted repeated experiments at the ``Small'' and ``Medium'' scales, performing five independent runs with different random seeds for each setting. As detailed in Table~\ref{tab:repeated-results}, DECO demonstrates highly stable and competitive performance. The narrow standard deviations confirm that DECO does not suffer from high variance across runs. Furthermore, DECO not only tightly matches the Dense baseline on the ``Small'' scale, but it actually slightly outperforms it on the ``Medium'' scale, achieving a lower average perplexity ($27.53 \pm 0.276$) and a higher average task score ($39.13 \pm 0.217$). Together, these robust empirical results confirm that DECO consistently maintains comparable and occasionally superior performance to the Dense setting.

\section{Theoretical Justification for NormSiLU} \label{sec:theoretical-normsilu}

To theoretically demonstrate the rationality of NormSiLU, we consider the post-activation output of all experts. Let $\mathbf{W}_{up} \in \mathbb{R}^{d_h \times (N_e d_e)}$ denote the concatenated up-projection weights of $N_e$ experts, and let $\mathbf{x} \in \mathbb{R}^{d_h}$ be the input hidden state. The post-activation intermediate state is:
\begin{equation}
\label{eq:output-normsilu}
    \mathbf{y}=\mathrm{SiLU}(\mathrm{Norm}(\mathbf{z})),\ \mathbf{z}=\mathbf{W}_{up}^T\mathbf{x} \in \mathbb{R}^{N_e d_e}.
\end{equation}
For simplicity, we first assume that the operator ``$\mathrm{Norm}$'' is implemented as a vanilla layer normalization~\cite{ba2016layer} across the entire concatenated expert dimension. Let $\mathbf{g} = \nabla_{\mathbf{y}}\mathcal{L}$ be the upstream gradient from the language modeling loss $\mathcal{L}$. The gradient with respect to the weights $\mathbf{W}_{up}$ is formulated as:
\begin{equation}
\begin{aligned}
\label{eq:gradient-normsilu}
    &\nabla_{\mathbf{W}_{up}}\mathcal{L} = \mathbf{x} (\nabla_{\mathbf{z}}\mathcal{L})^T = \mathbf{x} (\mathbf{J}_{norm}^T \mathbf{D}_{silu} \mathbf{g})^T, \\
    &\ \mathbf{u} = \mathrm{Norm}(\mathbf{z})=\frac{\mathbf{z_0}}{||\mathbf{z}_0||_{rms}},\ \mathbf{z}_0=\mathbf{z}-\bar{\mathbf{z}},
\end{aligned}
\end{equation}
where $\mathbf{D}_{silu} = \mathrm{diag}(\sigma(\mathbf{u})+\mathbf{u}\odot\sigma(\mathbf{u})\odot(1-\sigma(\mathbf{u})))$ is the diagonal Jacobian matrix of SiLU, $\sigma$ is the sigmoid function, and $\bar{\mathbf{z}}$ denotes the mean of $\mathbf{z}$. Letting $n = N_e d_e$ be the total dimension of $\mathbf{z}$ and $\mathbf{1}$ be the all-ones vector, the Jacobian matrix of the layer normalization $\mathbf{J}_{norm}$ is:
\begin{equation}
\begin{aligned}
\label{eq:gradient-layernorm}
    &\ \mathbf{J}_{norm} = \frac{\partial\mathbf{u}}{\partial\mathbf{z}} \\
    &= \frac{1}{||\mathbf{z}_0||_{rms}}\left(\mathbf{I}_n-\frac{1}{n}\mathbf{1}\mathbf{1}^T-\frac{\mathbf{z}_0\mathbf{z}_0^T}{n||\mathbf{z}_0||_{rms}^2}\right).
\end{aligned}
\end{equation}
Since the matrix inside the parentheses is an affine shift of a projection matrix, its spectral norm $||\mathbf{J}_{norm}||_2$ is strictly bounded by $1/||\mathbf{z}_0||_{rms}$. Furthermore, assuming the elements of $\mathbf{W}_{up}$ follow a zero-centered i.i.d. normal distribution, we can statistically approximate the RMS norm as:
\begin{equation}
    \label{eq:z0-rms-norm}
    ||\mathbf{z}_0||_{rms} \approx \frac{1}{\sqrt{d_h n}}||\mathbf{W}_{up}||_F \cdot ||\mathbf{x}||_2.
\end{equation}

Applying the sub-multiplicative property of matrix norms, the Frobenius norm of the gradient is bounded by:
\begin{equation}
\begin{aligned}
\label{eq:gradient-bound}
    &\ ||\nabla_{\mathbf{W}_{up}}\mathcal{L}||_F \\
    &\leq ||\mathbf{x}||_2 \cdot ||\mathbf{J}_{norm}||_2 \cdot ||\mathbf{D}_{silu}||_2 \cdot ||\mathbf{g}||_2 \\
    &\leq ||\mathbf{g}||_2 \cdot \mathcal{O}(||\mathbf{W}_{up}||_F^{-1}).
\end{aligned}
\end{equation}
Therefore, as long as $\mathbf{W}_{up}$ is initialized with a proper Frobenius norm, this normalization theoretically guarantees that the gradient scale remains bounded and invariant to the input magnitude, preventing gradient explosion.

However, the above paradigm possesses a critical systems-level flaw: computing a global layer normalization requires the explicit materialization of $\mathbf{z}$ across \textit{all} experts. This inherently violates the core sparsity principle of MoE, as it forces the computation of inactive experts. 

To resolve this bottleneck, our final implementation of NormSiLU decouples the operation into a dual-stage mechanism: an inter-expert mean normalization and an intra-expert RMS normalization. During inference, the global inter-expert averaging operator mathematically reduces to: $\mathrm{Avg}(\mathbf{W}_{up}^T\mathbf{x})=\bar{\mathbf{w}}_{up}^T\mathbf{x}$, where $\bar{\mathbf{w}}_{up} \in \mathbb{R}^{d_h\times d_e}$ is the fixed average up-projection weight. By contrast, the RMS normalization is strategically restricted to the intra-expert dimension ($d_e$) and applied \textit{only} to activated experts. This dual-stage design preserves the theoretical gradient-bounding stability while strictly adhering to the computational constraints of sparse inference.

\section{Evaluation Results on Individual Benchmarks} \label{sec:independent-benchmark}

In this section, we provide the evaluation results of DECO and baselines on each individual benchmark. The results of ``Small'', ``Medium'', ``Large'', and ``XLarge'' settings are shown in Table~\ref{tab:eval-small}, Table~\ref{tab:eval-medium}, Table~\ref{tab:eval-large}, and Table~\ref{tab:eval-xlarge}, respectively.

\begin{table*}[ht]
    \footnotesize
    \centering
    \begin{tabular}{l|ccccccc|c}
    \toprule
    & PIQA & SIQA & HellaSwag & ARC-E & ARC-C & WinoGrande & LAMBADA & \textbf{Avg.} \\
    \midrule
    TopP & 60.17 & 35.21 & 27.67 & 42.38 & 18.86 & 50.91 & 7.74 & 34.71 \\
    DeepSeek-V3 (GA) & 60.12 & 35.93 & 27.85 & 44.15 & 19.45 & 50.75 & 15.89 & 36.31 \\
    DeepSeek-V3 (NG) & 60.66 & 36.63 & 27.76 & 45.76 & 20.05 & 50.43 & 15.85 & 36.73 \\
    ReMoE (GA) & 60.34 & 36.59 & 27.72 & 46.34 & 18.43 & 51.22 & 15.60 & 36.61 \\
    ReMoE (NG) & 59.74 & 35.16 & 27.96 & 46.09 & 19.54 & 49.96 & 17.76 & 36.60 \\
    BlockFFN & 60.83 & 36.28 & 27.62 & 44.87 & 19.45 & 50.12 & 16.55 & 36.53 \\
    DECO (GA) & 58.22 & 35.93 & 27.65 & 42.51 & 19.03 & 51.30 & 15.84 & 35.78 \\
    \midrule
    \textbf{Dense} & 60.77 & 35.57 & 28.03 & 45.88 & 20.22 & 50.59 & 16.51 & \textbf{36.80} \\
    \textbf{DECO (NG)} & 60.94 & 36.85 & 27.96 & 45.84 & 20.22 & 50.67 & 15.80 & \textbf{36.90} \\
    \bottomrule
    \end{tabular}
    \caption{The evaluation scores of ``Small'' settings on individual benchmarks.}
    \label{tab:eval-small}
\end{table*}

\begin{table*}[ht]
    \footnotesize
    \centering
    \begin{tabular}{l|ccccccc|c}
    \toprule
    & PIQA & SIQA & HellaSwag & ARC-E & ARC-C & WinoGrande & LAMBADA & \textbf{Avg.} \\
    \midrule
    TopP & 61.37 & 36.49 & 28.57 & 45.75 & 19.03 & 50.51 & 12.63 & 36.34 \\
    DeepSeek-V3 (GA) & 62.79 & 37.87 & 29.61 & 48.53 & 20.90 & 52.17 & 19.84 & 38.82 \\
    DeepSeek-V3 (NG) & 63.28 & 37.67 & 29.64 & 49.07 & 20.65 & 49.25 & 19.76 & 38.47 \\
    ReMoE (GA) & 62.62 & 37.67 & 29.10 & 48.65 & 20.05 & 50.67 & 18.28 & 38.15 \\
    ReMoE (NG) & 63.22 & 36.64 & 29.40 & 49.71 & 19.71 & 50.83 & 20.45 & 38.57 \\
    BlockFFN & 63.38 & 35.82 & 29.61 & 48.82 & 20.99 & 50.51 & 19.95 & 38.44 \\
    DECO (GA) & 61.86 & 37.10 & 28.44 & 47.18 & 19.37 & 49.64 & 17.43 & 37.29 \\
    \midrule
    \textbf{Dense} & 63.00 & 36.90 & 29.57 & 51.47 & 22.01 & 50.51 & 19.62 & \textbf{39.01} \\
    \textbf{DECO (NG)} & 64.36 & 36.18 & 29.47 & 51.98 & 20.73 & 52.09 & 19.48 & \textbf{39.18} \\
    \bottomrule
    \end{tabular}
    \caption{The evaluation scores of ``Medium'' settings on individual benchmarks.}
    \label{tab:eval-medium}
\end{table*}

\begin{table*}[ht]
    \footnotesize
    \centering
    \begin{tabular}{l|ccccccc|c}
    \toprule
    & PIQA & SIQA & HellaSwag & ARC-E & ARC-C & WinoGrande & LAMBADA & \textbf{Avg.} \\
    \midrule
    TopP & 64.53 & 36.44 & 31.42 & 50.55 & 22.53 & 51.07 & 14.13 & 38.67 \\
    DeepSeek-V3 (GA) & 66.32 & 38.38 & 32.71 & 55.18 & 24.91 & 51.07 & 28.06 & 42.38 \\
    DeepSeek-V3 (NG) & 65.94 & 37.87 & 32.75 & 54.67 & 23.38 & 52.49 & 27.03 & 42.02 \\
    ReMoE (GA) & 66.05 & 38.69 & 32.08 & 55.26 & 24.32 & 52.33 & 27.19 & 42.27 \\
    ReMoE (NG) & 66.92 & 38.79 & 32.66 & 55.93 & 24.57 & 51.41 & 27.10 & 42.48 \\
    BlockFFN & 66.10 & 38.59 & 32.51 & 56.36 & 24.15 & 51.38 & 27.30 & 42.34 \\
    DECO (GA) & 66.14 & 38.74 & 32.69 & 56.41 & 23.72 & 51.78 & 26.57 & 42.29 \\
    \midrule
    \textbf{Dense} & 66.87 & 37.77 & 32.92 & 56.27 & 24.74 & 51.70 & 29.34 & \textbf{42.80} \\
    \textbf{DECO (NG)} & 65.94 & 39.61 & 32.78 & 57.11 & 25.43 & 53.28 & 25.50 & \textbf{42.81} \\
    \bottomrule
    \end{tabular}
    \caption{The evaluation scores of ``Large'' settings on individual benchmarks.}
    \label{tab:eval-large}
\end{table*}

\begin{table*}[ht]
    \footnotesize
    \centering
    \begin{tabular}{l|ccccccc|c}
    \toprule
    & PIQA & SIQA & HellaSwag & ARC-E & ARC-C & WinoGrande & LAMBADA & \textbf{Avg.} \\
    \midrule
    TopP & 66.16 & 37.36 & 34.10 & 57.32 & 22.87 & 49.41 & 20.49 & 41.10 \\
    DeepSeek-V3 (GA) & 69.53 & 39.92 & 36.65 & 62.21 & 28.41 & 53.28 & 34.52 & 46.36 \\
    ReMoE (NG) & 68.93 & 40.07 & 37.43 & 61.20 & 28.07 & 53.43 & 35.46 & 46.37 \\
    BlockFFN & 70.02 & 40.28 & 36.81 & 61.91 & 28.50 & 55.80 & 34.60 & 46.85 \\
    \midrule
    \textbf{Dense} & 70.46 & 40.69 & 37.33 & 63.38 & 28.58 & 52.72 & 35.73 & \textbf{46.98} \\
    \textbf{DECO (NG)} & 70.24 & 40.89 & 37.42 & 62.96 & 29.69 & 54.93 & 35.53 & \textbf{47.38} \\
    \bottomrule
    \end{tabular}
    \caption{The evaluation scores of ``XLarge'' settings on individual benchmarks.}
    \label{tab:eval-xlarge}
\end{table*}

\end{document}